\def\year{2020}\relax
\documentclass[letterpaper]{article} 
\usepackage{aaai20}  
\usepackage{times}  
\usepackage{helvet} 
\usepackage{courier}  
\usepackage[hyphens]{url}  
\usepackage{graphicx} 
\usepackage{adjustbox}
\usepackage{dashrule}
\usepackage{arydshln}
\usepackage{color}
\usepackage{amsmath}
\usepackage{amsfonts}
\usepackage{multirow}
\newcommand{\newcite}[1]{\citeauthor{#1}~\shortcite{#1}}
\newcommand{\citep}[1]{\citeauthor{#1},~\citeyear{#1}}
\usepackage{graphicx}  
\author{Anonymous Paper ID: 5400}
\title{DCR-Net: A Deep Co-Interactive Relation Network for Joint Dialog Act Recognition and Sentiment Classification
}
\author{Libo Qin,\textsuperscript{\rm }
	Wanxiang  Che,\textsuperscript{\rm }\thanks{Corresponding author}
	Yangming Li,\textsuperscript{\rm }
	Mingheng Ni,\textsuperscript{\rm }
	Ting Liu\textsuperscript{\rm } \\
	\textsuperscript{\rm }{Research Center for Social Computing and Information Retrieval}\\
	{Harbin Institute of Technology, Harbin, China}\\ \{lbqin, car, yangmingli, mhni, tliu\}@ir.hit.edu.cn
}

\begin{document}

\maketitle

\begin{abstract}
        In dialog system, dialog act recognition and sentiment classification are two correlative tasks to capture speakers’ intentions, where dialog act and sentiment can indicate the explicit and the implicit intentions separately  \cite{kim2018integrated}.
Most of the existing systems either treat them as separate tasks or just jointly model the two tasks by sharing parameters in an implicit way
without explicitly 
modeling mutual interaction and relation.
To address this problem, we propose a Deep Co-Interactive Relation Network (DCR-Net) to explicitly consider the cross-impact and model the interaction between the two tasks
by introducing a co-interactive relation layer. In addition, the proposed relation layer can be stacked to gradually capture mutual knowledge with multiple steps of interaction.
Especially, we thoroughly study different relation layers and their effects.
Experimental results on two public datasets (Mastodon and Dailydialog) show that 
our model outperforms the state-of-the-art joint model by 4.3\% and 3.4\% in terms of F1 score on dialog act recognition task, 5.7\% and 12.4\% on sentiment classification respectively.
Comprehensive analysis empirically verifies the effectiveness of explicitly modeling the relation between the two tasks and the multi-steps interaction mechanism.
Finally, we employ the Bidirectional Encoder 
Representation from Transformer (BERT) in our framework, which can further boost our performance in both tasks. 

\end{abstract}

\section{Introduction}

A dialog system should correctly understand speakers’ utterances and respond in natural language.
Dialog act recognition (DAR) and sentiment classification are two correlative tasks to realize the former.
The goal of DAR is to attach semantic labels to each utterance in a dialog and identify the underlying intentions \cite{kim2011review}.
Meanwhile, sentiment classification can detect the sentiments which are implicated in utterances and can help to capture speakers’ intentions \cite{kim2018integrated}.

\begin{figure}[t]
	\centering
	\includegraphics[scale=0.33]{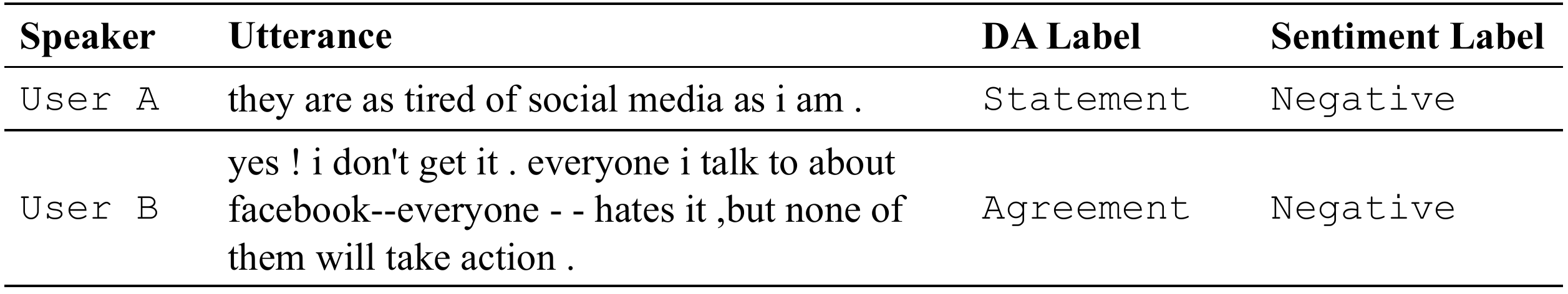}
	\caption{\label{corpus-examples}  A snippet of a dialog sample from the Mastodon Corpus and each utterance has a corresponding DA label and a sentiment label. (DA represents Dialog Act) }
\label{example1}
\end{figure}

Intuitively, the two tasks are closely related and the information of one task can be utilized in the other task.
For example, as illustrated in Figure~\ref{corpus-examples}, 
when predicting \texttt{User B} sentiment label, it's more likely to be  \texttt{Negative} in the case of known \texttt{Agreement} DA label, since \texttt{Agreement} means 
the current utterance agrees with previous \texttt{User A} utterance and hence \texttt{User B} sentiment label tends to be the same with the \texttt{User A} response sentiment \texttt{Negative}. 
Similarly, knowing the sentiment information also contributes to the current DA prediction.
Hence, it's promising to take the cross-impact between the two tasks into account.
In recent years, \citep{mastodon} has explored the multi-task framework to model the correlation between sentiment classification and dialog act recognition.
Unfortunately, their work does not achieve the promising performance, even underperforms some works which consider them as separate tasks.
In this paper, we argue that this modeling method with no explicit interaction between the two tasks is not effective enough for transferring knowledge across the two tasks and has following weaknesses: (1) A simple multi-task learning framework just implicitly considers mutual connection between two tasks by sharing latent representations, which cannot achieve desirable results \cite{ijcai2019-296}.
(2) With the shared latent representations, it is hard to explicitly control knowledge transfer for both tasks, resulting in lack of interpretability.

To address the aforementioned issues, we propose a {\bf{D}}eep \textbf{C}o-Interactive \textbf{R}elation \textbf{N}etwork (\textbf{DCR-Net}) for joint dialog act recognition
and sentiment classification, which can explicitly model relation and interaction between two tasks with a \textit{co-interactive relation layer}.
In practice, we first adopt a shared hierarchical encoder with utterance-level self-attention mechanism to obtain the shared representations of dialog act and sentiment among utterances.
The shared representations are then fed into the \textit{co-interactive relation layer} to get fusion of dialog act and sentiment representations and we call the process of fusion as one step
of interaction.
With the \textit{co-interactive relation layer}, we can directly control knowledge transfer for both tasks, which makes our framework more interpretable.
Besides,
the \textit{relation layer} can be stacked to form a hierarchy that enables multi-step interactions between the two tasks, which can further better capture mutual knowledge.
The underlying motivation is that if a model extracts mutual knowledge in one step of interaction, then by stacking multiple such steps, the model can gradually accumulate useful information and finally
 capture the semantic relation between the two tasks \cite{tao-etal-2019-one}. 
Specifically, we explore several \textit{relation layers} including:
1) \textit{Concatenation} that concats the representation of dialog act and sentiment.
2) \textit{Multilayer Perceptron (MLP)} that uses the \textit{MLP} to learn the rich representation which contains both dialog act and sentiment information.
3) \textit{Co-Attention} that uses the co-attention mechanism \cite{xiong2016dynamic} to capture mutually important information to contribute to the two tasks (sentiment to act and act to sentiment).
Finally, the final integrated outputs are then fed to separate decoders for dialog act and sentiment prediction respectively.

We conduct experiments on two real-world benchmarks including
Mastodon dataset \cite{mastodon} and Dailydialog dataset \cite{li-etal-2017-dailydialog}. The experimental results show that our system achieves significant 
and consistent improvement as compared to all baseline
methods and achieves the state-of-the-art performance.
Finally, Bidirectional Encoder Representation from Transformer (\citep{devlin-etal-2019-bert}, BERT), a pre-trained model, is used to further boost the performance.

To summarize, the contributions of this work are as follows:
\begin{itemize}
	\item We propose a deep co-interactive relation network for joint dialog act recognition and sentiment classification
	, which can explicitly control the cross knowledge transfer for both tasks and make our framework more interpretable.

	\item Our \textit{relation layer} can be stacked to form a hierarchy for 
	multi-step interactions between the two tasks, which can gradually capture mutual relation and better transfer knowledge.
	\item We thoroughly study different relation layers and present extensive experiments demonstrating the benefit of our proposed framework. Experiments on two publicly available datasets show substantial improvement and our framework achieves the state-of-the-art performance.
	
	\item Finally, we analyze the effect of incorporating BERT in our framework. With BERT, our framework reaches a new state-of-the-art level. 
\end{itemize}

\begin{figure*} [t]
	\centering
	\includegraphics[scale=0.9]{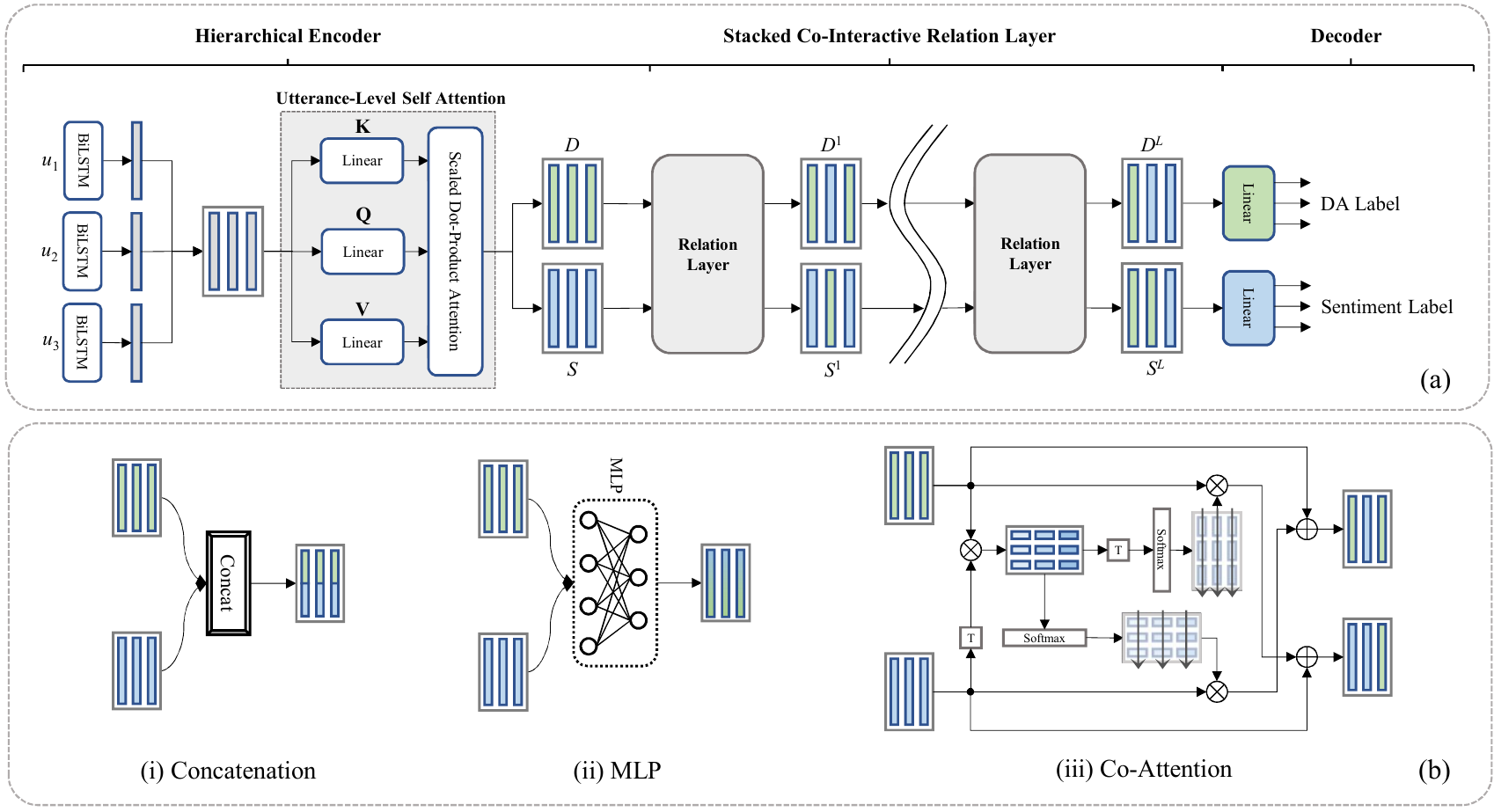}
	\caption{The top part (a) illustrates the overflow of our framework and the bottom part (b) represents different relation layers.
	}
	\label{fig:framework}
\end{figure*}

\section{Problem Formulation}
In this section, we describe the formulation definition for dialog act recognition and sentiment classification in dialog.
\begin{itemize}
	\item \textbf{Dialog Act Recognition} Given a dialog $C$ = $(u_{1},u_{2}...,u_{T})$ consisting of a sequence of $T$ utterances, dialog act recognition can be seen as a utterance-level sequence classification problem to decide the corresponding utterance dialog act label ($y_{1}^{d},y_{2}^{d},...,y_{T}^{d}$) for each utterance in dialog. 
	
	\item \textbf{Sentiment Classification in Dialog }  Sentiment classification in dialog can also be treated as an utterance-level sequence classification task that maps the utterance sequence $(u_{1},u_{2}...,u_{T})$ to the corresponding utterance sequence sentiment label ($y_{1}^{s},y_{2}^{s},...,y_{T}^{s}$).
\end{itemize}

\section{Our Approach}
In this section, we describe the architecture of DCR-Net; see the top part (a) of Figure.~\ref{fig:framework} for its overview.
DCR-Net mainly consists of three components: a shared hierarchical encoder, a stack of \textit{co-interactive relation layers} that repeatedly fuse dialog act and sentiment representations to explicitly model the relation and interaction between the two tasks, and two separate decoders for dialog act and sentiment prediction.
In the following sections, the details of our framework are given.

\subsection{Hierarchical Encoder}
In our framework, dialog act recognition and sentiment classification share one hierarchical encoder that consists of a bidirectional LSTM (BiLSTM) \cite{hochreiter1997long}, which captures temporal relationships within the words, followed by a utterance-level self-attention layer to consider the dialog contextual information.
\subsubsection{Utterance Encoder with BiLSTM}
Given a dialog $C$ = $(u_{1},...,u_{T})$ consists of a sequence of $T$ utterances and $u_{t}$ = $(w_{t}^{1},..., w_{t}^{K_{t}})$ which consists of a sequence of $K_{t}$ words,
we first adopt the BiLSTM to encode each utterance $u_{t}$$\in$$C$ to produce a series of hidden states (${ \mathbf { h } } _ t^1,..., { \mathbf { h } } _ t^{K_{t}}$), and we define ${\bf{h}}_{t}^{i}$ as follows:
\def\concat{\mathop{\rm concat}}
\begin{equation}
	\mathbf { h }_t^i = \concat{ \left( \overrightarrow { \mathbf { h } } _ t^i , \overleftarrow  { \mathbf { h } } _ t^i \right)},
\end{equation}
where $\concat(\cdot,\cdot)$ is an operation for concatenating two vectors, and $\overrightarrow { \mathbf { h } } _ t^i$ and $\overleftarrow  { \mathbf { h } } _ t^i$ are the $i$-th hidden state of the forward LSTM and backward LSTM for $w_t^i$ respectively. 

Then, we regard the last hidden state $ \mathbf{ h }_t^{K_{t}}$ as the utterance $u_{t}$ representation.
Hence, the sequence of $T$ utterances in $C$ can be represented as $\textbf{H}$ = ( $ \mathbf{ h }_1^{K_{1}}$, \dots, $ \mathbf{ h }_T^{K_{t}}$). 
\subsubsection{Utterance-Level Self-Attention}
Self-attention is an effective 
method of leveraging context-aware features over 
variable-length sequences for natural language 
processing tasks \cite{yin2017chinese,tan2018deep}. 
In our case, we use  self-attention mechanism to capture dialog-level
contextual information for each utterance.
In this paper, we adopt the self-attention formulation by \newcite{NIPS2017_7181} .
We first map the matrix of input vectors $\textbf{H}$ $\in$ $\mathbb{R}^{T\times d}$ 
($d$ represents the 
mapped dimension) to queries (${\textbf{Q}}$), keys (${\textbf{K}}$) and values 
(${\textbf{V}}$) matrices by different linear projections: 
\begin{equation}
\left[ \begin{array} { c } {  \textbf{K} } \\ { \textbf{Q} } \\ {\textbf{V} } \end{array} \right] = \left[ \begin{array} { c } { \textbf{W} _ { k } \textbf{H} } \\ { \textbf{W} _ { q } \textbf{H} } \\ { \textbf{W} _ { v } \textbf{H} } \end{array} \right].
\end{equation}

The attention weight is then computed by dot product 
between  $\textbf{Q}$, $\textbf{K}$ and the self-attention output 
{\bf{C}} $\in$ $\mathbb{R}^{T\times d}$  is a weighted sum of values $\textbf{V}$:
\begin{equation}
{{\textbf{C}}} = \text { Attention } ( \textbf{Q} ,\textbf{ K} , \textbf{V }) = \operatorname { Softmax } \left( \frac { \textbf{Q K }^ { T } } { \sqrt { d _ { k } } } \right) \textbf{V},
\end{equation}
where we can see ${\textbf{C}}$ = $(\mathbf { c }_1, ...,  \mathbf { c }_T)$ as the sequence utterances representations and each utterance representation captures the whole dialogue history information. $d_{k}$ represents the dimension of keys.

Now, we obtain the initial shared representations of sequence utterances dialog act ${\textbf{D}}$  = $( { \textbf{c} }_1, ...,   { \textbf{c} }_T)$ and sentiment representations $\bf{S}$ = $({ \textbf{c} }_1, ...,  { \textbf{c} }_T)$.

\subsection{Stacked Co-Interactive Relation Layer}
We now describe the proposed \textit{co-interactive relation layer}; see the bottom part (b) of Figure.\ref{fig:framework}.
In our paper, we use the \textit{co-interactive relation layer} to explicitly model the relation and interaction between dialog act recognition and sentiment classification.
It takes the dialog act ${\textbf{D}}$ and sentiment representations ${\textbf{S}}$ as inputs and then outputs their updated versions which consider cross-impact on two tasks.
In particular, it can be stacked to perform multi-step interaction for better capturing mutual knowledge and relation.
In our framework, we explore several types of relation layers, which can either be used individually or combined together.
Formally, given the $l^\text{th}$ layer inputs $\textbf{D}^{l}$ = ($\textbf{d}_{1}^{l}$, ...,$\textbf{d}_{T}^{l}$) $\in$ $\mathbb{R}^{T\times d}$  and  $\textbf{S}^{l}$ = ($\textbf{s}_{1}^{l}$, ...,$\textbf{s}_{T}^{l}$) $\in$ $\mathbb{R}^{T\times d}$, we can adopt the following strategies to integrate the mutual knowledge between the two tasks. Before fusing information, we first apply a BiLSTM and MLP over act information and sentiment information separately to make them more task-specific, which can be written as $\textbf{S}^{l^{'}}$ = MLP ($\textbf{S}^{l}$ ) and $\textbf{D}^{l^{'}}$ = BiLSTM ($\textbf{D}^{l}$ ).

{\bf{{Concatenation}}} Concatenation is an simple and effective method to combine two information \cite{wu2018improving}. Hence, we concatenate the $l^\text{th}$ layer of dialog act and sentiment representations as the updated representations.

\def\Concat{\mathop{\rm Concat}}
\begin{equation}
{ \textbf{D}}^{l+1} = \Concat ({ \textbf{S}}^{l^{'}},{ \textbf{D}}^{l^{'}}),
\end{equation}
\begin{equation}
{ \textbf{S}}^{l+1} = \Concat ({ \textbf{S}}^{l^{'}}, { \textbf{D}}^{l^{'}}).
\end{equation}

{\bf{{MLP}}} \textit{Multilayer Perceptron (MLP)} can automatically abstract the integrated representation \cite{nguyen2018improved}. Here, we add an MLP layer on the concatenation output to further learn the relation between two tasks and capture the mutual information, which can be formulated as follows:
\def\MLP{\mathop{\rm MLP}}
\def\Concat{\mathop{\rm Concat}}
\begin{equation}
{ \textbf{D}}^{l+1}  = \MLP (\Concat ({ \textbf{S}}^{l^{'}},{ \textbf{D}}^{l^{'}})) ,
\end{equation}
\begin{equation}
{ \textbf{S}}^{l+1}  = \MLP (\Concat ({ \textbf{S}}^{l^{'}},{ \textbf{D}}^{l^{'}})).
\end{equation}

{\bf{{Co-Attention}}} Co-Attention is a very effective method to grasp
the mutually important information on both correlated tasks \cite{xiong2016dynamic}.
Here, we extend the basic co-attention mechanism to utterance-level co-attention.
It can produce the updated dialog act representations considering sentiment information, and the updated
sentiment representations incorporating act knowledge.
By doing this, we can transfer mutually relevant knowledge for the two tasks.
The process can defined as follows:
\def\Softmax{\mathop{\rm Softmax}}
\def\Concat{\mathop{\rm Concat}}
\begin{equation}
{ \textbf{D}}^{l+1}= 	{ \textbf{D}}^{l^{'}} + \Softmax ({ \textbf{D}}^{l^{'}}( ({ \textbf{S}}^{l^{'}}) ^\top)) { \textbf{S}}^{l^{'}},
\end{equation}
\begin{equation}
{ \textbf{S}}^{l+1} = { \textbf{S}}^{l^{'}} + \Softmax ({ \textbf{S}}^{l^{'}}( ({ \textbf{D}}^{l^{'}}) ^\top)){ \textbf{D}}^{l^{'}},
\end{equation}
where ${\textbf{D}}^{l+1}$ = ($\textbf{d}_{1}^{l+1}$, ...,$\textbf{d}_{T}^{l+1}$) and  ${\textbf{S}}^{l+1}$ = ($\textbf{s}_{1}^{l+1}$, ...,$\textbf{s}_{T}^{l+1}$) are the $l^\text{th}$ layer updated representations.

\subsection{Decoder for Dialog Act Recognition and Sentiment Classification}
After multi-step interaction with stacked co-interactive relation layer, we can get the outputs ${\textbf{D}}^{L}$ = ($\textbf{d}_{1}^{L}$, ...,$\textbf{d}_{T}^{L}$)  and $\textbf{S}^{L}$ = ($\textbf{s}_{1}^{L}$, ...,$\textbf{s}_{T}^{L}$) of the last relation layer. We then adopt separate decoder to perform dialog act and sentiment prediction, which can be denoted as follows: 
\def\softmax{\mathop{\rm softmax}}
\begin{equation}
	\textbf{y}_{t}^{d} = \softmax (\textbf{W}^{d}{ \textbf{d} }_{t}^{L} + \textbf{b}_{d}),
\end{equation}
\begin{equation}
\textbf{y}_{t}^{s} = \softmax (\textbf{W}^{s}{\textbf{s} }_{t}^{L} + \textbf{b}_{s}),
\end{equation}
where $\textbf{y}_{t}^{d} $ and $\textbf{y}_{t}^{s}$ are the predicted distribution for dialog act and sentiment respectively; $\textbf{W}^{d}$ and $\textbf{W}^{s}$ are transformation matrices; $\textbf{b}_{d}$ and $\textbf{b}_{s}$ are bias vectors; $L$ is the number of stacked relation layers in our framework.
\subsection{Joint Training}
The dialog act recognition objection is formulated as:

\begin{equation}
	\mathcal { L } _ { 1 } = - \sum _ { i = 1 } ^ {  T}  \hat { {\bf{y}} } _ { i } ^ { d } \log \left( {\bf{y}} _ { i} ^ {  d } \right).
\end{equation}
Similarly, the sentiment classification objection is defined as: 
\begin{equation}
	\mathcal { L } _ { 2 } = - \sum _ { i = 1 } ^ {  T}  \hat { {\bf{y}} } _ { i } ^ { s } \log \left( {\bf{y}} _ { i} ^ { s} \right),
\end{equation}
where ${\hat { {\bf{y}} } _ { i } ^ { d } }$ and 
$ {\hat { {\bf{y}}} _ { i } ^ { s } }$ are gold utterance act label and 
gold sentiment label separately.

To obtain dialog act recognition and sentiment classification jointly, we follow \newcite{qin-etal-2019-stack} to obtain the final joint objective:
\begin{equation}
	\mathcal { L } _ { \theta } = \mathcal{ L }_{1} + \mathcal{ L }_{2}.
\end{equation}

\section{Experiments}
\subsection{Dataset} \label{sec:dataset}
We evaluate the performance of our model on two publicly available dialogue datasets, Mastodon \cite{mastodon} and Dailydialog \cite{li-etal-2017-dailydialog}.

{\bf{Mastodon}} The Mastodon dataset\footnote{https://github.com/cerisara/DialogSentimentMastodon} consists of 269 dialogues
for a total of 1075 utterances in training dataset and the test dataset is a corpus of 266 dialogues for a total of 1142 utterances. The vocabulary size is 5330.
We follow the same partition as \newcite{mastodon}.

{\bf{DailyDialog}} For Dailydialog dataset,\footnote{http://yanran.li/dailydialog} we adopt the standard split from the original dataset \cite{li-etal-2017-dailydialog}, employing 11,118 dialogues for training, 1,000 for validating, and 1,000 for testing.

\begin{table*}[th]
	\small
	\centering
	\begin{adjustbox}{width=0.9\textwidth}
		\begin{tabular}{l|ccc|ccc|ccc|ccc} 
			\hline 
			\multirow{3}*{\textbf{Model}} & \multicolumn{6}{c}{\textbf{Mastodon}} & \multicolumn{6}{c}{\textbf{Dailydialog}} \\ 
			\cline{2-13} 
			~ & \multicolumn{3}{c|}{SC} & \multicolumn{3}{c|}{DAR} & \multicolumn{3}{c|}{SC} & \multicolumn{3}{c}{DAR} \\ 
			\cline{2-13}  
			~ & F1 (\%) & R (\%) & P (\%) & F1 (\%) & R (\%) & P (\%) & F1 (\%) & R (\%) & P (\%) & F1 (\%) & R (\%) & P (\%) \\  
			\hline

			HEC \cite{kumar2018dialogue} & - &-   & -&56.1  &55.7  &56.5 &-  &- &- &77.8  &76.5  &77.8  \\  
			
			CRF-ASN \cite{chen2018dialogue} & - & -  &- &55.1  &53.9  &56.5 &-  &- & -&76.0  &75.6  &78.2 \\ 
			
			CASA \cite{raheja-tetreault-2019-dialogue} & - &-   &- &56.4  &57.1  &55.7 & - &- &- &78.0  &76.5  &77.9 \\ 
			\hline 
			
			VDCNN \cite{conneau-etal-2017-deep} &39.6 & 31.6& 44.0 & - & - & - & 39.7 & 35.6 & 55.2 & - & - & - \\  
			
			Region.emb \cite{qiao2018new} &40.3 & 33.6 & 42.8 & - & - & - & 41.0 & 36.6 & 56.4 & - & - & -\\ 
			
			DRNN \cite{wang-2018-disconnected} &37.9 & 34.3 & 39.7 & - & - & - & 41.1 & 37.0 & 56.4 & - & - & - \\  

						DialogueRNN \cite{majumder2019dialoguernn} &41.5 &42.8 & 40.5 & - & - & - & 40.3 & 37.7 & 44.5 & - & - & - \\  
			\hline 
			JointDAS \cite{mastodon} & 37.6 & 41.6 & 36.1 & 53.2 & 51.9 & 55.6 & 31.2 & 28.8 & 35.4 & 75.1 & 74.5 & 76.2 \\  
			
			IIIM \cite{kim2018integrated} & 39.4 & 40.1 & 38.7 & 54.3 & 52.2 & 56.3 & 33.0 & 28.5 & 38.9 & 75.7 & 74.9 & 76.5 \\ 
			\hline 
			DCR-Net + Concat &42.1 & 41.3 & 42.9 & 57.1 & 56.9 & 57.2 & 41.2 & 37.4 & 57.4 & 78.2 & 77.6 & 78.7 \\  
			
			DCR-Net + MLP & 42.3 & 43.7 & 45.4 & 57.2 & 56.7 & 57.7 & 42.7 & 37.5 & \textbf{58.8} & 79.1 & 78.5 &{ 79.2} \\ 
			
			DCR-Net + Co-Attention & \bf{*45.1} & \bf{*47.3} & \bf{*43.2} & \bf{*58.6} & \bf{*56.9} & \bf{*60.3} & \bf{*45.4} & \bf{*40.1} & 56.0 & \bf{*79.1} & \bf{*79.0} & \bf{*79.1} \\  
						\hdashline  
				DCR-Net + Co-Attention + BERT & 55.1 & 56.5 & 56.5 & 67.1 & 65.2 & 69.2 & 48.9 & 46.9 & 63.2 & 80.0 & 79.9 & 80.2 \\  
			\hline  

		\end{tabular}
	\end{adjustbox}
	\caption{Comparison of our model with baselines on Mastodon and Dailydialog test datasets. SC represents Sentiment Classification and DAR represents Dialog Act Recognition. The numbers with * indicate that the improvement of our model over all baselines is statistically significant with $p<0.01$ under t-test.}
	\label{table:over_all}
	
\end{table*} 
\subsection{Experimental Settings}
In our experiment setting,  dimensionality of the embedding and all hidden units is selected from $\{100, 128, 256, 512, 600, 700, 800, 1024\}$.
We do not use any pre-trained embedding and all word embeddings are trained from scratch.
L2 regularization used on our model is $1\times 10^{-8}$ and dropout ratio adopted is selected from $\{0.1, 0.2, 0.25, 0.3, 0.4, 0.5\}$.
In addition, we add a residual connection in self-attention and relation layer for reducing overfitting.
We use Adam \cite{kingma-ba:2014:ICLR} to 
optimize the parameters in our model and adopt
the suggested hyper-parameters for optimization.
We set the stacked number of relation layer as 3.
For all experiments, we pick the model which works best on dev set, and then evaluate it on test set.

\subsection{Baselines}
We first make a comparison with the state-of-the-art dialog act recognition models:  HEC, CRF-ASN and CASA, and then we compare our model with some state-of-the-art sentiment classification models: VDCNN, Region.emb, DRNN and DialogueRNN. Finally, we compare our framework with the existing state-of-the-art joint models including: JointDAS and IIIM.
We briefly describe these  baseline models below: 1) \textbf{{HEC}} \cite{kumar2018dialogue}: This work uses a hierarchical Bi-LSTM-CRF (Bi-directional Long Short Term Memory with CRF) model for dialog act recognition, which can capture both kinds of dependencies including word-level and utterance-level. 2) \textbf{{CRF-ASN}} \cite{chen2018dialogue}: This model proposes a crf-attentive structured network for dialog act recognition, which can dynamically separate the utterances into cliques. 3) \textbf{{CASA}} \cite{raheja-tetreault-2019-dialogue}: This work leverages a context-aware self-attention mechanism coupled with a hierarchical deep neural network and achieves state-of-the-art performance. 4) \textbf{{VDCNN}} \cite{conneau-etal-2017-deep}: This work proposes a deep CNN with 29 convolutional layers for text classification. 5) \textbf{{Region.emb}} \cite{qiao2018new}: This work proposes a new method of region embedding for text classification, which can effectively learn and utilize task-specific distributed representations of n-grams. 6) \textbf{{DRNN}} \cite{wang-2018-disconnected}: This work proposes a disconnected recurrent neural network for 
	text classification which can incorporate position-invariance into RNN. 7) \textbf{{DialogueRNN}} \cite{majumder2019dialoguernn}: DialogueRNN is a RNN-based neural architecture for emotion detection in a conversation, which can keep track of the individual party states throughout the conversation and uses this information. 8) \textbf{{JointDAS}} \cite{mastodon}: This model uses a multi-task modeling framework for joint dialog act recognition and sentiment classification, which models relation and interaction between two tasks by sharing parameters. 9) \textbf{{IIIM}} \cite{kim2018integrated}: This work proposes an integrated neural network model which simultaneously identifies speech acts, predicators, and sentiments of dialogue utterances.
	
For \textit{HEC, CRF-ASN, CASA} and \textit{IIM} we re-implemented the models.
For \textit{VDCNN, Region.emb}, \textit{DRNN} and \textit{DialogueRNN}, we adopted the open-sourced code\footnote{https://github.com/Tencent/NeuralNLP-NeuralClassifier and https://github.com/senticnet/conv-emotion} to get the results.
For \textit{JointDAS}, we adopted the reported results from \newcite{mastodon} and run their open-source code on Dailydialog dataset to obtain results.
For \textit{IIIM}, we re-implemented the model and obtained results on the same datasets.\footnote{All experiments are conducted on the public datasets provided by \newcite{mastodon} and the dataset does not annotate the predictors. For direct comparison, we re-implemented the models excepting predicting the predicators and obtained the results on the same dataset.} For all BERT-based experiments, we just replace our utterance encoder LSTM with BERT base model.\footnote{The BERT model is fine-tuned with our framework.}

\subsection{Overall Results}
On Dailydialog dataset, following \newcite{kim2018integrated},
we adopt macro-average Precision, Recall and F1 for both sentiment classification and dialog act recognition.
On Mastodon dataset,
following \newcite{mastodon}, we ignore the neural sentiment label and adopt the average of the dialog-act specific F1 scores weighted by the prevalence of each dialog act for dar.
The experimental result is shown in Table~\ref{table:over_all}.
From the result, we can observe that: 
\begin{enumerate}
	\item We obtain large improvements compared with prior joint models. In Mastodon dataset, compared with \textit{IIIM} model, our framework with Co-Attention achieves 5.7\% improvement on F1 score on sentiment classification task and 4.3\% improvement on F1 score on dialog act recognition task.
	In Dailydialog dataset, 
	we achieve 12.4\% improvement on F1 score on sentiment classification task and 3.4\% improvement on F1 score on dialog act recognition task.
	It is worth noting that the prior joint models have modeled the relation between two tasks implicitly by sharing parameters. This result demonstrates the effectiveness of explicitly modeling the interaction between the two tasks and both tasks can boost performance from this mechanism.
	\item Our framework with Co-Attention outperforms the state-of-the-art dialog act recognition models and sentiment classification models in all metrics in two datasets.
	It illustrates the advantages and effectiveness of our proposed joint model where the information of one task can be effectively utilized in the other task.
	\item The \textit{MLP} relation layer outperforms the \textit{concatenation}, which shows that the  \textit{MLP} can further learn the deep implicit relation
	between two tasks and improve the performance. Especially, we can see that the \textit{Co-Attention} relation layer gains the best performance among three relation layers on F1 scores on all datasets. 
	We attribute this to the fact that the \textit{Co-Attention} operation can automatically detect the mutually important information to each other and better interact with the two tasks.
	 \item{From the last block of Table~\ref{table:over_all}, the BERT-based model performs remarkably well on both two datasets and achieves a new state-of-the-art performance, which indicates the effectiveness of a strong pre-trained model in two tasks. We attribute this to the fact that pre-trained models can provide rich semantic features, which can help to improve the performance. 
 }
\end{enumerate}
Unless otherwise stated, we only apply the Co-Attention relation layer in the following experiments.
\subsection{Analysis} \label{sec:analysis}
Although achieving good performance, we would like to know the reason for the improvement. In this section, we study our model from several directions. 
We first conduct several ablation experiments to analyze the effect of different components in our framework. 
Next, we give a quantitative analysis to study how our proposed framework improves performance.
Finally, we provide a co-attention visualization to better understand how relation layer affects and contributes to the performance.

\subsubsection{Ablation}
In this section, we perform several ablation experiments in our framework on two datasets and the results are shown in Table~\ref{table:no_relation} .
The results demonstrate the effectiveness of different components of our framework to the final performance.
We give a detailed analysis in the following:
\begin{itemize}
	\item \textbf{w/o relation layer:} In this settings, we conduct experiments on the multi-task framework where dialog act recognition and sentiment classification promote each other only by sharing parameters of the encoder, 
	which similar to \newcite{mastodon}. 
	From the result, we can see that 4.8\% drop in terms of F1 scores in sentiment classification while 2.6\% drops in dialog act recognition in Mastodon dataset.
	In Dailydialog dataset, we can also observe the same trends that the F1 score drops a lot. 
	This demonstrates that explicitly modeling the strong relations between two tasks with relation layer can 
	benefit them effectively.
	
	\begin{table}[t]
		
		\centering
		\begin{adjustbox}{width=0.45\textwidth}
			\begin{tabular}{l|cc|cc} 
				\hline 
				
				\multirow{2}*{Model} & \multicolumn{2}{c|}{Mastodon} & \multicolumn{2}{c}{Dailydialog} \\ 
				\cline{2-5}        
				
				~ & SC (F1) & DAR (F1) & SC (F1) & DAR (F1) \\ 
				\hline  
				
				Full Model & 45.1 & 58.6 & 45.4 & 79.1 \\
				\hline
				w/o relation layer & 40.3 & 55.2 & 38.0 & 78.4 \\
			
				w/o stacked relation layer & 42.5 & 57.4 & 42.1 & 78.5 \\
					w/o self-attention & 43.2 & 57.3 & 42.1 & 77.2 \\
						\ \ \ \ +CNN & 43.9 &	58.2 &	43.1 &	78.4 \\
					
				\hline
				
			\end{tabular}
		\end{adjustbox}
		\caption{Ablation study on Mastodon and Dailydialog test datasets.}
		\label{table:no_relation}
	\end{table} 

	\item \textbf{w/o stacked relation layer:} Here, we set the number of the stacked relation layer as 1 in our framework. From the result, we can see that performance drops in all metrics.
It indicates that stacked structure with multiple steps of interaction does better 
model the semantic relation.

	\item \textbf{w/o self-attention:} In this setting, we remove our self-attention layer and there is no hierarchical architecture to capture dialog-level context information.
	The results show a significant drop in
	performance, indicating that capturing the dialog-level context information by the hierarchical encoder is effective and important for dialog act recognition and sentiment classification. 
	In addition, we replace our self-attention with CNN \cite{kim-2014-convolutional} which can also model the dialog context information. The result is shown in the last row of Table~\ref{table:no_relation}.
We can see that CNN outperforms \textit{w/o self-attention} version and underperforms our full model, which further demonstrates the effectiveness of the dialog context information and self-attention mechanism.
	
\end{itemize}

\subsubsection{Quantitative Analysis}
\begin{figure}[t]
	\centering
	\includegraphics[scale=0.5]{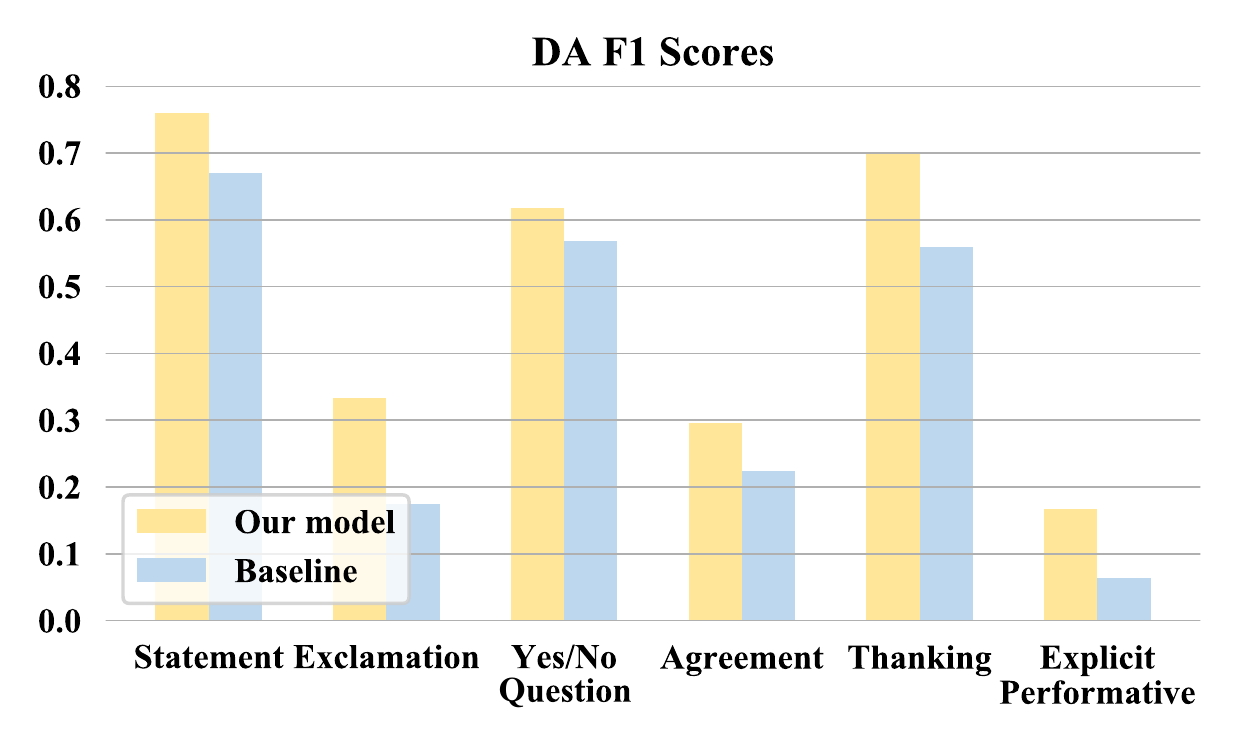}
	\caption{Quantitative analysis on different types of DA between our model with baseline.}
	\label{fig:DA}
\end{figure}

\begin{figure}[t]
	\centering
	\includegraphics[scale=0.4]{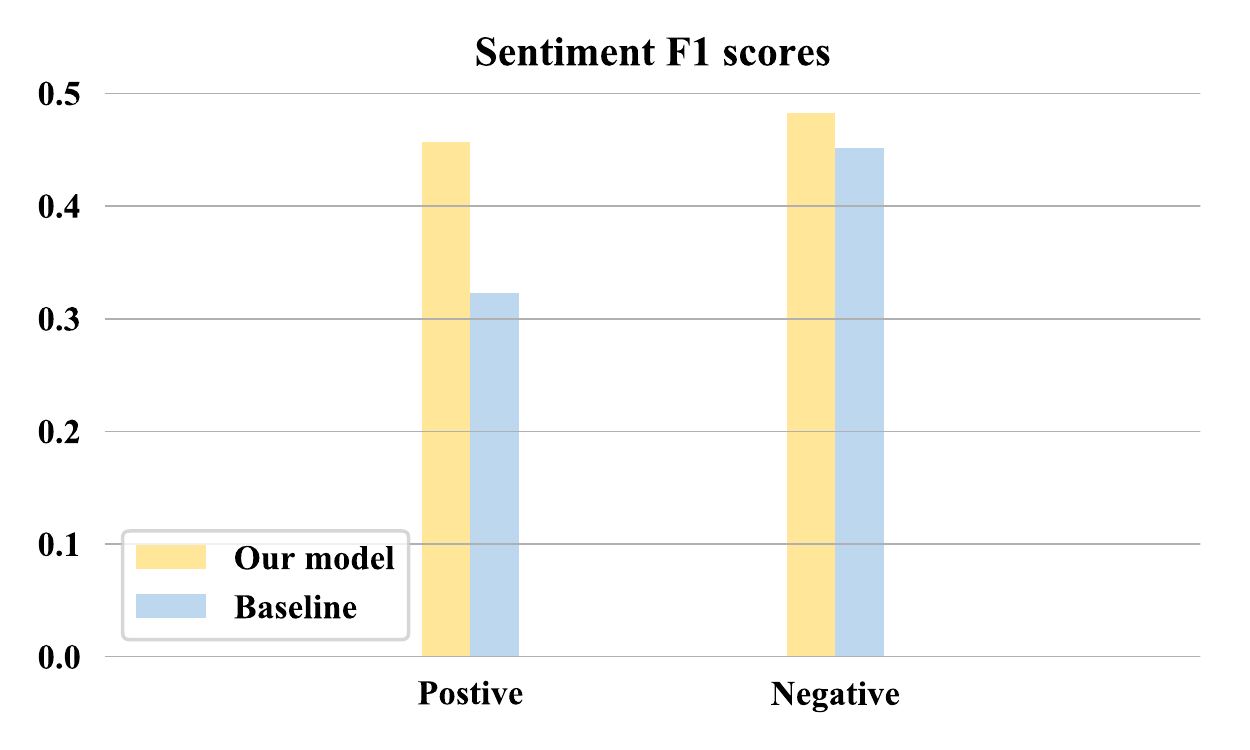}
	\caption{Quantitative analysis on different types of sentiments between our model with baseline.}
	\label{fig:sentiment}
\end{figure}

In our DCR-Net model, we adopt the relation layer to model the interaction and relation between two tasks explicitly.
To better understand our model, we compare the DA and sentiment performance between DCR-Net
model and baseline without relation layer, as shown in Figure~\ref{fig:DA} and Figure~\ref{fig:sentiment}.

\begin{figure*}[t]\label{visualization}
	\centering
	\includegraphics[scale=0.5]{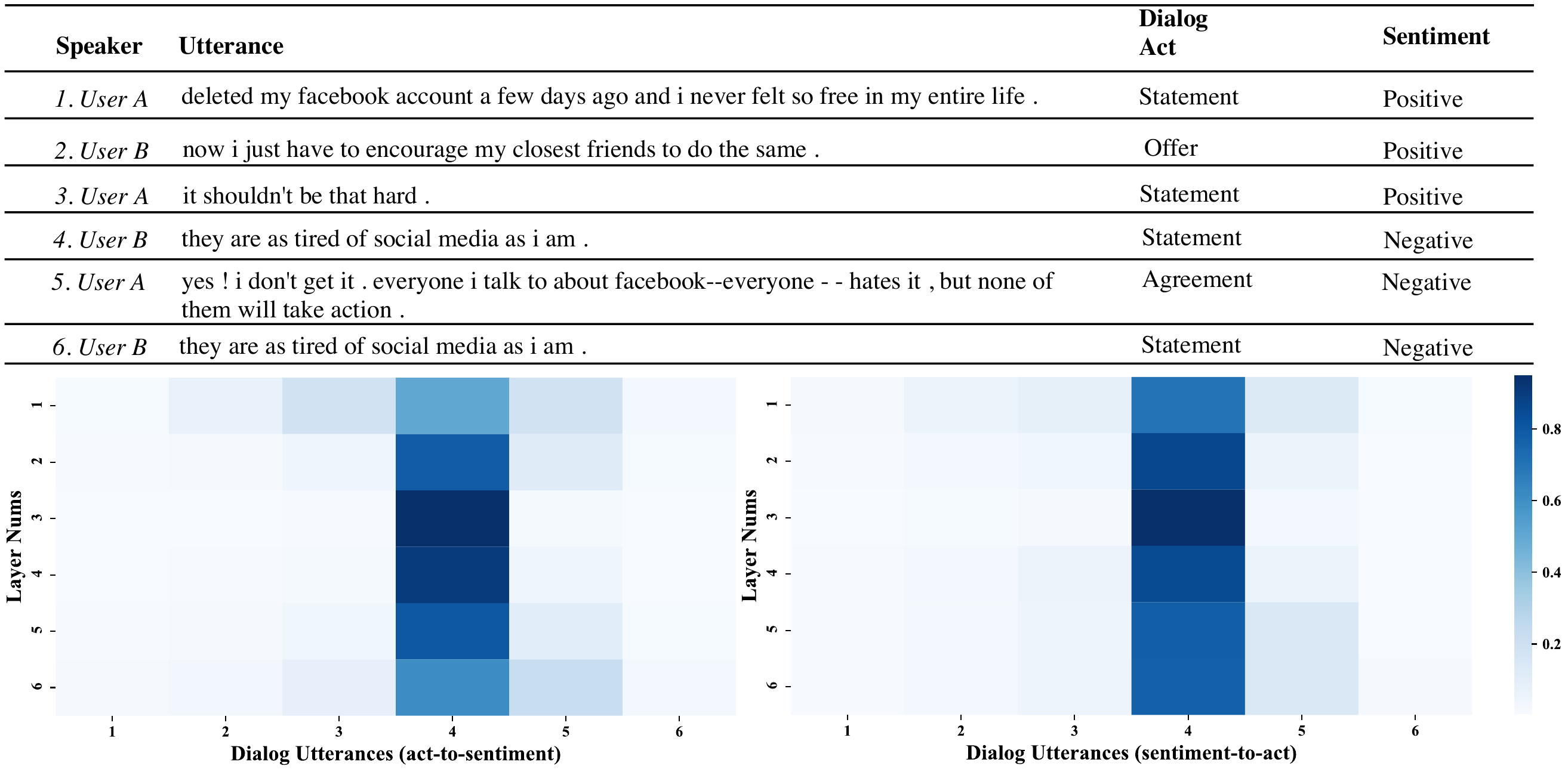}
	\caption{Co-Attention distribution score from the fifth utterance to the whole dialog utterances. The top is corresponding dialog context. The bottom left is the act-to-sentiment attention and the right is the sentiment-to-act attention. }
	\label{fig:visual}
\end{figure*}

We choose several DA types with a large performance boost which are shown in Figure~\ref{fig:DA}.
From the results, we can see that our model yields significant improvements on the act type \texttt{Exclamation}, \texttt{Thanking}, \texttt{Agreement}, \texttt{Explicit Performative}.
We attribute the improvements to the fact that those acts are strong correlative with sentiment and our model can provide sentiment information explicitly for DAR rather than in an implicit method by sharing parameters. 
Take the fourth utterance in Figure~\ref{fig:visual} for example, providing the current utterance \texttt{Negative} sentiment information explicitly and previous utterance sentiment \texttt{Negative} label can contribute to DA \texttt{Agreement} prediction, which demonstrates the effectiveness of  our proposed framework.
In addition, from Figure~\ref{fig:sentiment}, we can observe that our model outperforms baseline in both positive and negative sentiment label.
We think that our relation layer can explicitly capture DA information which benefits sentiment classification task.
	
\subsection{Visualization of Co-Attention}
In this section, with the attempt to better understand what the model has learnt,
we visualized
the co-attention distribution among utterances in dialogues.
In particular, 
we visualized the attention distribution of the fifth utterance corresponding to other utterances with the number of stacked relation layers varying from 1 to 6.

From Figure~\ref{fig:visual}, we can observe: 
(1) the act-to-sentiment attention distribution score in the fourth utterance is larger than other utterances. This is due to that the fifth utterance is more related to the fourth utterance and the \texttt{Agreement} DA represents that current utterance agrees with the fourth utterance statement.
Similarly, we can see that sentiment-to-act attention    
in the fourth utterance distribution score is also the largest compared to other utterances.
Those results demonstrate that our framework can correctly capture mutually important knowledge. 
(2) Using deeper layers could generally lead to better performance, especially when the number of stacked layers is less than four. It is because the stacked relation layer can better model the relation between two tasks and learn mutual knowledge. When the number of stacked layers exceeds three, the experimental performance goes worse. We suggest that the reason might lie in the gradient vanishing or overfitting problem as the whole network goes deeper.

\section{Related Work}
In this section, we will introduce the related work about dialog act recognition, sentiment classification and the joint model for the two tasks. 
\subsection{Dialog Act Recognition}
Recently, more and more neural networks have been propose to solve the DAR. \newcite{kalchbrenner2013recurrent} propose the hierarchical CNN to model the utterance sequence for DA classification.
\newcite{lee2016sequential} propose a model based on CNNs and RNNs which incorporated the previous utterance as context to classify the current DA and show the promising performance.
\newcite{ji2016latent} propose the latent variable recurrent neural network for jointly modeling sequences of words and discourse relations between adjacent sentences.
Furthermore, many work \cite{liu2017using,kumar2018dialogue,chen2018dialogue} explore different architectures to incorporate the context information for DAR.
\newcite{raheja-tetreault-2019-dialogue} propose the token-level self-attention mechanism for DAR and achieved state-of-the-art performance.
\subsection{Sentiment Classification}
Sentiment classification in dialog system can be seen as the sentence-level sequence classification problem.
One series of works are based on CNN \cite{zhang2015character,conneau-etal-2017-deep,johnson-zhang-2017-deep} to capture the local correlation and position-invariance.
Another series of works adopt RNN based models \cite{tang-etal-2015-document,yang-etal-2016-hierarchical,xu-etal-2016-cached} to leverage temporal features and contextual information to perform sentence classification.
Besides, Some works \cite{xiao2016efficient,shi-etal-2016-deep,wang-2018-disconnected} attempt to combine the advantages of CNN and RNN for sentence classification.
\subsection{Joint Model}
Considering the correlation between dialog act recognition and sentiment classification, joint models are proposed to solve two tasks simultaneously in a unified framework.
\newcite{mastodon} explore the multi-task framework to model the correlation between the two tasks.
Compared with their model, we propose a relation layer to explicitly model the correlation between dialog act recognition and sentiment classification while they model in an implicit way simply by sharing parameters. 
Specifically, our relation layer can be stacked to capture mutual knowledge sufficiently.
\newcite{kim2018integrated} propose an integrated 
neural network model for identifying dialog act, predicators, and sentiments of dialogue utterances.
Their framework classifies the current dialog act only considering the last time dialog act results, which can not make full use of context information, while we adopt the hierarchical encoder with utterance-level self-attention  to leverage context information. 
In addition, their model does not model the sentiment information for dialog act while our framework considers interaction and mutual relation between two tasks.

\section{Conclusion}
This paper focuses on explicitly establishing the bi-directional interrelated connections for dialog act recognition and sentiment information.
We propose a deep relation network to jointly model the interaction and relation between the two tasks, which adopts a stacked co-interactive relation layer to incorporate mutual knowledge explicitly.
In addition, we explore three different relation layers and make a thorough study on their effects on the two tasks.
Experiments on two datasets show the effectiveness of the proposed models and achieve the state-of-the-art performance. 
Extensive analysis further confirms the correlation between two tasks and reveals that modeling the relation explicitly can boost their performance.
Besides, we analyze the effect of incorporating strong pre-trained BERT model in our joint model. With BERT, the result reaches a new state-of-the-art level.

\section{Acknowledgments}
We thank the anonymous reviewers for their helpful comments and suggestions.
This work was supported by the National Natural Science Foundation of China (NSFC) via grant 61976072, 61632011 and 61772153.
	\bibliography{5400-References.bib}

\begin{thebibliography}{}

\bibitem[\protect\citeauthoryear{Cerisara \bgroup et al\mbox.\egroup
  }{2018}]{mastodon}
Cerisara, C.; Jafaritazehjani, S.; Oluokun, A.; and Le, H.~T.
\newblock 2018.
\newblock Multi-task dialog act and sentiment recognition on mastodon.
\newblock In {\em Proc. of COLING}.

\bibitem[\protect\citeauthoryear{Chen \bgroup et al\mbox.\egroup
  }{2018}]{chen2018dialogue}
Chen, Z.; Yang, R.; Zhao, Z.; Cai, D.; and He, X.
\newblock 2018.
\newblock Dialogue act recognition via crf-attentive structured network.
\newblock In {\em Proc. of SIGIR}.

\bibitem[\protect\citeauthoryear{Chen \bgroup et al\mbox.\egroup
  }{2019}]{ijcai2019-296}
Chen, Z.; Wang, X.; Xie, X.; Wu, T.; Bu, G.; Wang, Y.; and Chen, E.
\newblock 2019.
\newblock Co-attentive multi-task learning for explainable recommendation.
\newblock In {\em Proc. of IJCAI}.

\bibitem[\protect\citeauthoryear{Conneau \bgroup et al\mbox.\egroup
  }{2017}]{conneau-etal-2017-deep}
Conneau, A.; Schwenk, H.; Barrault, L.; and Lecun, Y.
\newblock 2017.
\newblock Very deep convolutional networks for text classification.
\newblock In {\em Proc. of ACL}.

\bibitem[\protect\citeauthoryear{Devlin \bgroup et al\mbox.\egroup
  }{2019}]{devlin-etal-2019-bert}
Devlin, J.; Chang, M.-W.; Lee, K.; and Toutanova, K.
\newblock 2019.
\newblock {BERT}: Pre-training of deep bidirectional transformers for language
  understanding.
\newblock In {\em Proc. of NAACL}.

\bibitem[\protect\citeauthoryear{Hochreiter and
  Schmidhuber}{1997}]{hochreiter1997long}
Hochreiter, S., and Schmidhuber, J.
\newblock 1997.
\newblock Long short-term memory.
\newblock {\em Neural computation}.

\bibitem[\protect\citeauthoryear{Ji, Haffari, and
  Eisenstein}{2016}]{ji2016latent}
Ji, Y.; Haffari, G.; and Eisenstein, J.
\newblock 2016.
\newblock A latent variable recurrent neural network for discourse relation
  language models.
\newblock {\em arXiv preprint arXiv:1603.01913}.

\bibitem[\protect\citeauthoryear{Johnson and
  Zhang}{2017}]{johnson-zhang-2017-deep}
Johnson, R., and Zhang, T.
\newblock 2017.
\newblock Deep pyramid convolutional neural networks for text categorization.
\newblock In {\em Proc. of ACL}.

\bibitem[\protect\citeauthoryear{Kalchbrenner and
  Blunsom}{2013}]{kalchbrenner2013recurrent}
Kalchbrenner, N., and Blunsom, P.
\newblock 2013.
\newblock Recurrent convolutional neural networks for discourse
  compositionality.
\newblock {\em arXiv preprint arXiv:1306.3584}.

\bibitem[\protect\citeauthoryear{Kim and Kim}{2018}]{kim2018integrated}
Kim, M., and Kim, H.
\newblock 2018.
\newblock Integrated neural network model for identifying speech acts,
  predicators, and sentiments of dialogue utterances.
\newblock {\em Pattern Recognition Letters}.

\bibitem[\protect\citeauthoryear{Kim, Seon, and Seo}{2011}]{kim2011review}
Kim, H.-S.; Seon, C.-N.; and Seo, J.-Y.
\newblock 2011.
\newblock Review of korean speech act classification: machine learning methods.
\newblock {\em Journal of Computing Science and Engineering}.

\bibitem[\protect\citeauthoryear{Kim}{2014}]{kim-2014-convolutional}
Kim, Y.
\newblock 2014.
\newblock Convolutional neural networks for sentence classification.
\newblock In {\em Proc. of EMNLP}.

\bibitem[\protect\citeauthoryear{Kingma and Ba}{2014}]{kingma-ba:2014:ICLR}
Kingma, D.~P., and Ba, J.
\newblock 2014.
\newblock Adam: A method for stochastic optimization.
\newblock {\em arXiv preprint arXiv:1412.6980}.

\bibitem[\protect\citeauthoryear{Kumar \bgroup et al\mbox.\egroup
  }{2018}]{kumar2018dialogue}
Kumar, H.; Agarwal, A.; Dasgupta, R.; and Joshi, S.
\newblock 2018.
\newblock Dialogue act sequence labeling using hierarchical encoder with crf.
\newblock In {\em Proc. of AAAI}.

\bibitem[\protect\citeauthoryear{Lee and Dernoncourt}{2016}]{lee2016sequential}
Lee, J.~Y., and Dernoncourt, F.
\newblock 2016.
\newblock Sequential short-text classification with recurrent and convolutional
  neural networks.
\newblock {\em arXiv preprint arXiv:1603.03827}.

\bibitem[\protect\citeauthoryear{Li \bgroup et al\mbox.\egroup
  }{2017}]{li-etal-2017-dailydialog}
Li, Y.; Su, H.; Shen, X.; Li, W.; Cao, Z.; and Niu, S.
\newblock 2017.
\newblock Dailydialog: A manually labelled multi-turn dialogue dataset.
\newblock In {\em Proc. of IJCNLP}.

\bibitem[\protect\citeauthoryear{Liu \bgroup et al\mbox.\egroup
  }{2017}]{liu2017using}
Liu, Y.; Han, K.; Tan, Z.; and Lei, Y.
\newblock 2017.
\newblock Using context information for dialog act classification in dnn
  framework.
\newblock In {\em Proc. of EMNLP}.

\bibitem[\protect\citeauthoryear{Majumder \bgroup et al\mbox.\egroup
  }{2019}]{majumder2019dialoguernn}
Majumder, N.; Poria, S.; Hazarika, D.; Mihalcea, R.; Gelbukh, A.; and Cambria,
  E.
\newblock 2019.
\newblock Dialoguernn: An attentive rnn for emotion detection in conversations.
\newblock In {\em Proc. of AAAI}.

\bibitem[\protect\citeauthoryear{Nguyen and Okatani}{2018}]{nguyen2018improved}
Nguyen, D.-K., and Okatani, T.
\newblock 2018.
\newblock Improved fusion of visual and language representations by dense
  symmetric co-attention for visual question answering.
\newblock In {\em Proc. of CVPR}.

\bibitem[\protect\citeauthoryear{Qiao \bgroup et al\mbox.\egroup
  }{2018}]{qiao2018new}
Qiao, C.; Huang, B.; Niu, G.; Li, D.; Dong, D.; He, W.; Yu, D.; and Wu, H.
\newblock 2018.
\newblock A new method of region embedding for text classification.
\newblock In {\em Proc. of ICLR}.

\bibitem[\protect\citeauthoryear{Qin \bgroup et al\mbox.\egroup
  }{2019}]{qin-etal-2019-stack}
Qin, L.; Che, W.; Li, Y.; Wen, H.; and Liu, T.
\newblock 2019.
\newblock A stack-propagation framework with token-level intent detection for
  spoken language understanding.
\newblock In {\em Proc. of EMNLP}.

\bibitem[\protect\citeauthoryear{Raheja and
  Tetreault}{2019}]{raheja-tetreault-2019-dialogue}
Raheja, V., and Tetreault, J.
\newblock 2019.
\newblock {D}ialogue {A}ct {C}lassification with {C}ontext-{A}ware
  {S}elf-{A}ttention.
\newblock In {\em Proc. of NAACL}.

\bibitem[\protect\citeauthoryear{Shi \bgroup et al\mbox.\egroup
  }{2016}]{shi-etal-2016-deep}
Shi, Y.; Yao, K.; Tian, L.; and Jiang, D.
\newblock 2016.
\newblock Deep {LSTM} based feature mapping for query classification.
\newblock In {\em Proc. of NAACL}.

\bibitem[\protect\citeauthoryear{Tan \bgroup et al\mbox.\egroup
  }{2018}]{tan2018deep}
Tan, Z.; Wang, M.; Xie, J.; Chen, Y.; and Shi, X.
\newblock 2018.
\newblock Deep semantic role labeling with self-attention.
\newblock In {\em Proc. of AAAI}.

\bibitem[\protect\citeauthoryear{Tang, Qin, and
  Liu}{2015}]{tang-etal-2015-document}
Tang, D.; Qin, B.; and Liu, T.
\newblock 2015.
\newblock Document modeling with gated recurrent neural network for sentiment
  classification.
\newblock In {\em Proc. of ACL}.

\bibitem[\protect\citeauthoryear{Tao \bgroup et al\mbox.\egroup
  }{2019}]{tao-etal-2019-one}
Tao, C.; Wu, W.; Xu, C.; Hu, W.; Zhao, D.; and Yan, R.
\newblock 2019.
\newblock One time of interaction may not be enough: Go deep with an
  interaction-over-interaction network for response selection in dialogues.
\newblock In {\em Proc. of ACL}.

\bibitem[\protect\citeauthoryear{Vaswani \bgroup et al\mbox.\egroup
  }{2017}]{NIPS2017_7181}
Vaswani, A.; Shazeer, N.; Parmar, N.; Uszkoreit, J.; Jones, L.; Gomez, A.~N.;
  Kaiser, L.~u.; and Polosukhin, I.
\newblock 2017.
\newblock Attention is all you need.
\newblock In {\em Proc. of NIPS}. Curran Associates, Inc.

\bibitem[\protect\citeauthoryear{Wang}{2018}]{wang-2018-disconnected}
Wang, B.
\newblock 2018.
\newblock Disconnected recurrent neural networks for text categorization.
\newblock In {\em Proc. of ACL}.

\bibitem[\protect\citeauthoryear{Wu \bgroup et al\mbox.\egroup
  }{2018}]{wu2018improving}
Wu, Z.; Dai, X.-Y.; Yin, C.; Huang, S.; and Chen, J.
\newblock 2018.
\newblock Improving review representations with user attention and product
  attention for sentiment classification.
\newblock In {\em Proc. of AAAI}.

\bibitem[\protect\citeauthoryear{Xiao and Cho}{2016}]{xiao2016efficient}
Xiao, Y., and Cho, K.
\newblock 2016.
\newblock Efficient character-level document classification by combining
  convolution and recurrent layers.
\newblock {\em arXiv preprint arXiv:1602.00367}.

\bibitem[\protect\citeauthoryear{Xiong, Merity, and
  Socher}{2016}]{xiong2016dynamic}
Xiong, C.; Merity, S.; and Socher, R.
\newblock 2016.
\newblock Dynamic memory networks for visual and textual question answering.
\newblock In {\em Proc. of ICML}.

\bibitem[\protect\citeauthoryear{Xu \bgroup et al\mbox.\egroup
  }{2016}]{xu-etal-2016-cached}
Xu, J.; Chen, D.; Qiu, X.; and Huang, X.
\newblock 2016.
\newblock Cached long short-term memory neural networks for document-level
  sentiment classification.
\newblock In {\em Proc. of EMNLP}.

\bibitem[\protect\citeauthoryear{Yang \bgroup et al\mbox.\egroup
  }{2016}]{yang-etal-2016-hierarchical}
Yang, Z.; Yang, D.; Dyer, C.; He, X.; Smola, A.; and Hovy, E.
\newblock 2016.
\newblock Hierarchical attention networks for document classification.
\newblock In {\em Proc. of NAACL}.

\bibitem[\protect\citeauthoryear{Yin \bgroup et al\mbox.\egroup
  }{2017}]{yin2017chinese}
Yin, Q.; Zhang, Y.; Zhang, W.; and Liu, T.
\newblock 2017.
\newblock Chinese zero pronoun resolution with deep memory network.
\newblock In {\em Proc. of EMNLP},  1309--1318.

\bibitem[\protect\citeauthoryear{Zhang, Zhao, and
  LeCun}{2015}]{zhang2015character}
Zhang, X.; Zhao, J.; and LeCun, Y.
\newblock 2015.
\newblock Character-level convolutional networks for text classification.
\newblock In {\em Proc. of NIPS},  649--657.

\end{thebibliography}
	\bibliographystyle{aaai}
\end{document}